\documentclass[11pt,a4paper]{article}
\usepackage[hyperref]{acl2017}
\usepackage{times}
\usepackage{latexsym}

\usepackage{enumerate}
\usepackage{amsfonts, amsmath, amsthm, amssymb} 
\usepackage{multirow}
\usepackage{covington}
\usepackage{graphicx}
\usepackage{fixltx2e}
\usepackage{subfig}

\usepackage{url}

\aclfinalcopy 
\def\aclpaperid{157} 

\title{German in Flux:\\
Detecting Metaphoric Change via Word Entropy}

\author{Dominik Schlechtweg$^{*}$$^{\dag}$, Stefanie Eckmann$^{\ddag}$, Enrico Santus$^{\Diamond}$, Sabine Schulte im Walde$^{*}$, Daniel Hole$^{\dag}$ \\
	$^{*}$Inst. for Natural Language Processing, University of Stuttgart, Germany \\
	$^{\dag}$Dept. of Linguistics/German Studies, University of Stuttgart, Germany \\
	$^{\ddag}$Historical and Indo-European Linguistics, LMU Munich \\
	$^{\Diamond}$Singapore University of Technology and Design, Singapore \\
	{\normalsize {\tt dominik.schlechtweg@gmx.de}, {\tt stefanie.eckmann@campus.lmu.de}, {\tt esantus@mit.edu},} \\
	{\normalsize {\tt schulte@ims.uni-stuttgart.de}, {\tt holedan@gmail.com}}
}

\date{\today}

\begin{document}
\maketitle
\begin{abstract}
This paper explores the information-theoretic measure entropy to detect metaphoric change, transferring ideas from hypernym detection to research on language change. We also build the first diachronic test set for German as a standard for metaphoric change annotation. Our model shows high performance, is unsupervised, language-independent and generalizable to other processes of semantic change.
\end{abstract}


\section{Introduction}

Recently, computational linguistics has shown an increasing interest in language change. This interest is focused on making semantic change measurable. However, even though different types of semantic change are well-known in historical linguistics, little effort has been made to distinguish between them. A very basic distinction in historical linguistics is the one between \textit{innovative meaning change} (also polysemization)---e.g., German \textit{br\"uten} `breed' $>$ `breed, brood over sth.'---and \textit{reductive meaning change}---e.g., German \textit{schinden} `to skin, torture' $>$ `to torture' \cite[cf.][p.~24--27]{Koch:2016}. Metaphoric meaning change is an important sub-process of innovative meaning change. Hence, a computational model of semantic change should be able to distinguish metaphoric change from other---typically less strong---types of change. Such a model, particularly if applicable to different languages, would be beneficial for a number of areas: (i), historical linguists may test their theoretical claims about semantic change on a large-scale empirical basis going beyond the traditional corpus-based approaches; (ii), linguists and psychologists working on metaphor in language or cognition may benefit by gaining new insights into the diachronic aspects of metaphor which are not yet as central in these fields as the synchronic aspects; and, finally, (iii), the Natural Language Processing research community may benefit by applying the model presented here to a wide range of tasks in which polysemy and non-literalness are involved. 

Our aim is to build an unsupervised and language-independent computational model which is able to distinguish metaphoric change from semantic stability. We apply \textit{entropy} (a measure of uncertainty inherited from information theory) to a \textit{Distributional Semantic Model} (DSM). In particular, we exploit the idea of \textit{semantic generality} applied in hypernym detection, to detect metaphoric change as a special process of meaning innovation. German will serve as a sample language, since there is a rich historical corpus available covering a large time period. Nevertheless, our model is presumably applicable to other languages requiring only minor adjustments. With the model, we introduce the first resource for evaluation of models of metaphoric change and propose a structured annotation process that is generalizable to the creation of gold standards for other types of semantic change.\footnote{The test set is provided together with the annotation data and the model code (which is based on \citet[][]{Shwartz16}'s code): \url{https://github.com/Garrafao/MetaphoricChange}}

In the next section, we give an overview of related work on semantic change and automatic detection of metaphor. In Section \ref{sec:metaphoric}, the basic linguistic notions we focus on are introduced and connected to their distributional properties, followed by a description of the corpus used to obtain vector representations of words in Section \ref{sec:corpus}. In Section \ref{sec:entropy}, the information-theoretic measures we apply to word vectors are described. Section \ref{sec:annot} presents the annotation study conducted to create a metaphoric change test set for German. Section \ref{sec:evaluation} illustrates how the measures' predictions shall be evaluated. The results are presented and discussed in Section \ref{sec:Results}. Section \ref{sec:Conclusion} will then conclude and give a short outlook to further research objectives.

\section{Related Work}
\label{subsec:relatedwork}

There is a number of recent approaches to trace semantic change via distributional methods. This includes mainly (i), semantic \textit{similarity models} assuming one sense for each word and then measuring its spatial displacement by a similarity metric (such as cosine) in a semantic vector space \citep[][]{Gulordava11,Kim14,Xu15,Eger:2016,Hellrich16p2785,Hamilton16,Hamilton:2016} and (ii), \textit{word sense induction models} (WSI) inferring for each word a probability distribution over different word senses (or topics) in turn modeled as a distribution over words \citep[][]{Wang06,Bamman11p1,Wijaya11p35,Lau12p591,Mihalcea12,Frermann:2016}. 

Most of the similarity models seem to be limited to quantify the degree of overall change rather than being able to qualify different types of semantic change.\footnote{With the exception of \citet[][p.~1]{Hamilton16} making the rather coarse-grained distinction between cultural shift and ``regular processes of linguistic drift''.} Similarity metrics, in particular, were shown not to distinguish well between words on different levels of the semantic hierarchy \citep[][]{Shwartz16}. Thus, we cannot expect diachronic similarity models to reflect changes in the semantic generality of a word over time, which was described to be a central effect of semantic change \citep[cf.][p.~197]{Bybee:2015aa}. Additionally, they often pose the problem of vector space alignment (especially when relying on word embeddings), occurring when word vectors from different time periods have to be mapped to a common coordinate axis \citep[cf.][p.~1492]{Hamilton:2016}.

Diachronic WSI models, on the contrary, are able to detect at least innovative (and reductive) meaning change, as they are designed to induce newly arising senses of words. However, they do not measure how these senses relate to each other in terms of semantic generality. Hence, ad hoc, they may not be able to distinguish different subtypes of innovative meaning change such as metaphoric vs. metonymic change. They may fail to detect meaning changes where no new senses can be induced as, e.g., in grammaticalization. Moreover, some models require elaborate training \citep[e.g.,][]{Frermann:2016}. 

Apart from similarity and WSI models, \citet{Sagi09p104} measure semantic broadening and narrowing of words (shifting upwards and downwards in the semantic taxonomy respectively) via \textit{semantic density} calculated as the average cosine of its context word vectors. Just as word entropy, semantic density is based on the measurement of linguistic context dispersion (see Section \ref{sec:distributional}). However, this method is only applied in a case study with very limited scope in terms of the number of phenomena covered and there is no verification of the test items via annotation. Hence, it remains to be shown that the method can generally distinguish broadening and narrowing or other types of meaning innovation.

Two previous approaches to language change exploit the notion of entropy. \citet{Juola:2003} describes language change on a very general level by computing the relative entropy (or KL-divergence) of language stages, i.e. intuitively speaking, measuring how well later stages of English encode a prior stage. \citet{kisselewetal:2016} are interested in the diachronic properties of conversion using---among other measures---a word entropy measure.

Finally, research on synchronic metaphor identification has applied a wide range of approaches, including binary classification relying on standard distributional similarity \cite{Birke/Sarkar:06}, text cohesion measures \cite{Li/Sporleder:09}, classification relying on abstractness cues \cite{TurneyEtAl:11,Koeper/SchulteImWalde:16b} or cross-lingual information \cite{TsvetkovEtAl:14}, and soft clustering \cite{ShutovaEtAl:13}, among others. As to our knowledge, no previous work has explicitly exploited the idea of generalization (via hypernymy models) in metaphor detection yet.


\section{Metaphoric Change}
\label{sec:metaphoric}

Metaphoric change plays a fundamental role in semantic change \citep[cf.~e.g.][p.~15]{Ferraresi:2014}. Within the framework of Conceptual Metaphor Theory \citep[][]{Lakoff80} the metaphorical effect can be described as a mapping from a source domain to a target domain. 
Following the terminology from \citet[][p.~24]{Koch:2016} \textit{innovative meaning change}, as opposed to \textit{reductive meaning change}, is where the existing meaning $M_A$ (the source concept) of a word acquires a new meaning $M_B$ (the target concept). \textit{Metaphoric Change} is, then, a subcategory of innovative meaning change where $M_B$ is related to $M_A$ by similarity or a reduced comparison (\citealp[cf.][p.~47]{Koch:2016}, \citealp[and also][p.~10]{Steen10}). 
While language is often used ad hoc in a non-literal meaning in discourse, not every of these uses constitutes an instance of metaphoric change. Only when a metaphoric innovation is conventionalized within the language, we can speak of metaphoric meaning change \citep[cf.][p.~27]{Koch:2016}. 
Consider German \textit{umw\"alzen} as an example. In Early New High German the word was only used in the sense `to turn around something or someone physically' ($M_A$) as in (\ref{ex:1}).\footnote{Early New High German: ca. 1350-1650; Contemporary New High German: 1650-today \citep[cf.][p.~24]{Fleischer:2011}} In Contemporary New High German, though, the word is also frequently used in the sense `to change something (possibly abstract) radically' ($M_B$) as in (\ref{ex:2}).

\begin{examples}
\item \label{ex:1} {\em ...mu{\ss} ich mich \textbf{vmbweltzen} / vnd kan keinen schlaff in meine augen bringen} \footnote{Neomenius, J.: Christliche Leichpredigt. Brieg, 1616.}
      \glt `...I have to turn around and cannot bring sleep into my eyes.'
\item \label{ex:2} {\em Kinadon wollte den Staat \textbf{umw\"alzen}...} \footnote{M\"uller, K. O.: Die Dorier. Vier B\"ucher. Bd. 2, 1824.}
      \glt `Kinadon wanted to revolutionize the state...'
\end{examples}


\subsection{Distributional Properties}
\label{sec:distributional}

As \citet{Bybee:2015aa} notes, and is also commonly agreed-upon, ``metaphorical meaning changes create \textit{polysemy}'' (p.~199, her italics). \citet[][p.~258]{campbell1998historical} describes this effect as ``extensions in the meaning of a word'' occurring through metaphoric change. It is only logical to assume that such extensions in meaning range imply an extension in the range of linguistic contexts a word occurs in. This extension, then, distinguishes words undergoing such a change from semantically stable words, but also from words undergoing different types of meaning change such as reductive meaning change where we expect an oppositional effect: a reduction of the range of contexts a word occurs in. Polysemization (and thus context extension) is, yet, not only a typical property of metaphoric change but of all types of innovative meaning change such as \textit{metonymic change}, \textit{generalization}, \textit{specialization}, and \textit{grammaticalization} \citep[cf.][p.~35]{heineKut:2007}.
However, recall that metaphor involves a mapping between two different domains (as introduced in \citealt[][]{Lakoff80}) in contrast to other types of meaning change, which is why we would expect a relatively strong effect on the contextual distribution here. 

Moreover, not only the range of a word's meanings influences the range of contexts it occurs in, but also the particular nature of the individual meanings has an influence. As research in hypernymy detection shows, words at different levels of semantic generality have different distributional properties \citep[][]{Rimell14,Santus:2014,Shwartz16}. According to the \textit{distributional informativeness hypothesis}, semantically more general words are less informative than special words as they occur in more general contexts \citep[][]{Rimell14,Santus:2014}. Hence, differences in \textit{semantic generality} of source and target concept should be reflected by their contextual distribution.\footnote{Related ideas are also indicated, e.g., by \citet[p.~650]{Fortson:2003aa} and \citet[p.~202]{Bybee:2015aa}.} Such differences occur particularly with taxonomic meaning changes like generalization and specialization, but also with metaphoric change, as it often results in the emergence of more abstract meanings of a word.
Consider, e.g., the development of German \textit{gl\"anzend} with `luminous' as source and `very good' as target concept. The source concept only applies to a rather limited range of entities, i.e., physical ones. The target concept, on the contrary, given its abstractness, applies to nearly every entity. Interpreting such changes of words as a change in their semantic generality, we now aim to examine how well it is measurable with distributional methods.


\section{Corpus}
\label{sec:corpus}

For our investigation, we use the corpus of \textit{Deutsches Textarchiv (erweitert)} (DTA), which is accessible online and downloadable for free.\footnote{\url{http://www.deutschestextarchiv.de/}} The DTA provides more than 2447 lemmatized and POS-tagged texts (with more than 140M tokens), covering a time period from the late 15\textsuperscript{th} to the early 20\textsuperscript{th} century. Thus, it covers the developments of German from (late) Early New High German to Contemporary New High German. The corpus is POS-tagged using the STTS tagset \citep{stts_tagset:1999}. The texts used by DTA include literary and scientific texts as well as functional writings, e.g., cookbooks. DTA aims at providing a corpus with a roughly equivalent number of texts from each of the aforementioned genres. The corpus is preprocessed in standard ways. (Find details in Appendix \ref{sec:params}.) For the creation of the co-occurrence matrices, from which we calculate word entropy and the other measures, a standard model of distributional semantics with a symmetric window of size 2 is used.

\section{Entropy}
\label{sec:entropy}

In hypernym detection a number of well-established measures compare the semantic generality of words on the basis of their distributional generality \citep{Weeds:2003,clarke:2009,Kotlerman:2009}. A promising candidate measure seems to be \textit{word entropy}, which is introduced in \citet{santus2013slqs} and \citet{Santus:2014}. Amongst other advantages, word entropy is independently measurable over time, which avoids the problem of vector space alignment.

\subsection{Entropy in Information Theory}

The term `Entropy' was first introduced by \citet{Shannon:1948} who laid the foundations of information theory. Intuitively, it measures the unpredictability of a system.
The entropy $H$ of a discrete random variable $X$ with possible values $\{x_1, ..., x_n\}$ and probability mass function $P(X)$ (a probability distribution) is

\begin{equation}
\label{eq:entropy}
H(X) = -\sum_{i=1}^{n} P(x_i) \log_bP(x_i)
\end{equation} where $b$ is typically equal to 2 or 10 \cite[cf.~p.~11]{Shannon:1948}. 

\paragraph{Word Entropy.} Examining language statistically, a word $w$ may be represented by its distribution in a corpus. This distribution is determined by the contexts of $w$, i.e., the words it co-occurs with, and how often it co-occurs with them. The distribution of $w$ is usually recorded in a matrix, intuitively a table where rows correspond to target word distributions and columns to context word distributions. Rows are typically referred to as \textit{vectors} and the whole matrix spans a \textit{vector space}. We can interpret $w$'s (normalized) vector then as a probability distribution where word co-occurrences of $w$ with any other corpus word $w'$ correspond to events in the probability distribution. More specifically, assuming that $C$ and $T$ are discrete random variables of occurrences of context and target words respectively, we say that $w$'s vector estimates the conditional probability distribution of context words given target word $w$ with discrete random variable $C$ and a probability mass function defined by $P(C\mid T = w)$. For every $c \in C$, $P(c\mid w)$ (the probability that the context word $c$ will occur given the occurrence of $w$ as target word) is estimated by $\frac{Freq(w,c)}{Freq(w)}$.\footnote{For convenience, here, we do not distinguish between a word and the mathematical structure corresponding to the event of the occurrence of the word.} Now, we can apply any notion from probability theory to this distribution. Hence, the entropy of $w$'s probability distribution is given by
{\begin{equation}
H(C) = -\sum_{i=1}^{n} P(c_i\mid w) \log_2 P(c_i\mid w)
\end{equation}}The entropy of $w$'s estimated probability distribution---for the sake of convenience we will just write $H(w)$---measures the unpredictability of $w$'s co-occurrences, i.e., how hard it is to predict with which word $w$ will co-occur if we look at a random occurrence of $w$. In hypernym detection, word entropy is assumed to reflect semantic generality. While here it is mostly used to compare pairs of different words for their semantic relations, e.g., whether one is the hypernym of the other, we will compare the word entropy of one and the same word $w$ in different time periods assuming this to reflect $w$'s semantic development with respect to its generality.

\subsubsection*{Normalization}

Depending on corpus size and other factors, the frequency of each target word will vary strongly. On top of that, the number of types in the corpus increases with the progression of time. These factors influence word entropy (and also other measures) without being tied to semantic change. Hence, we need a way to normalize for them. We test essentially two ways of normalizing word entropy for word frequency:

\paragraph{Matching Occurrence Number (MON).} The first strategy assumes that, for the most part, the influence of word frequency on word entropy comes from the increasing number of context types with increasing number of contexts $n$ used to construct a word vector (where $n$ is dependent on word frequency). Hence, we can suppress the influence of word frequency by comparing only word vectors constructed from an equal number of contexts \citep[cf.][]{kisselewetal:2016}. In order to make the vectors of all target words from all time periods comparable, we choose a common number of contexts $n$ for all target words. Additionally, in order to diminish the influence of chance (because we do not use all contexts, we have to pick a random subset), we average over the entropies computed for a number of $k$ vectors, each constructed from a different $n$-sized set of contexts. (Find information on the setting of hyperparameters in Appendix \ref{sec:params}.) 

\begin{figure}
\includegraphics[width=0.48\textwidth]{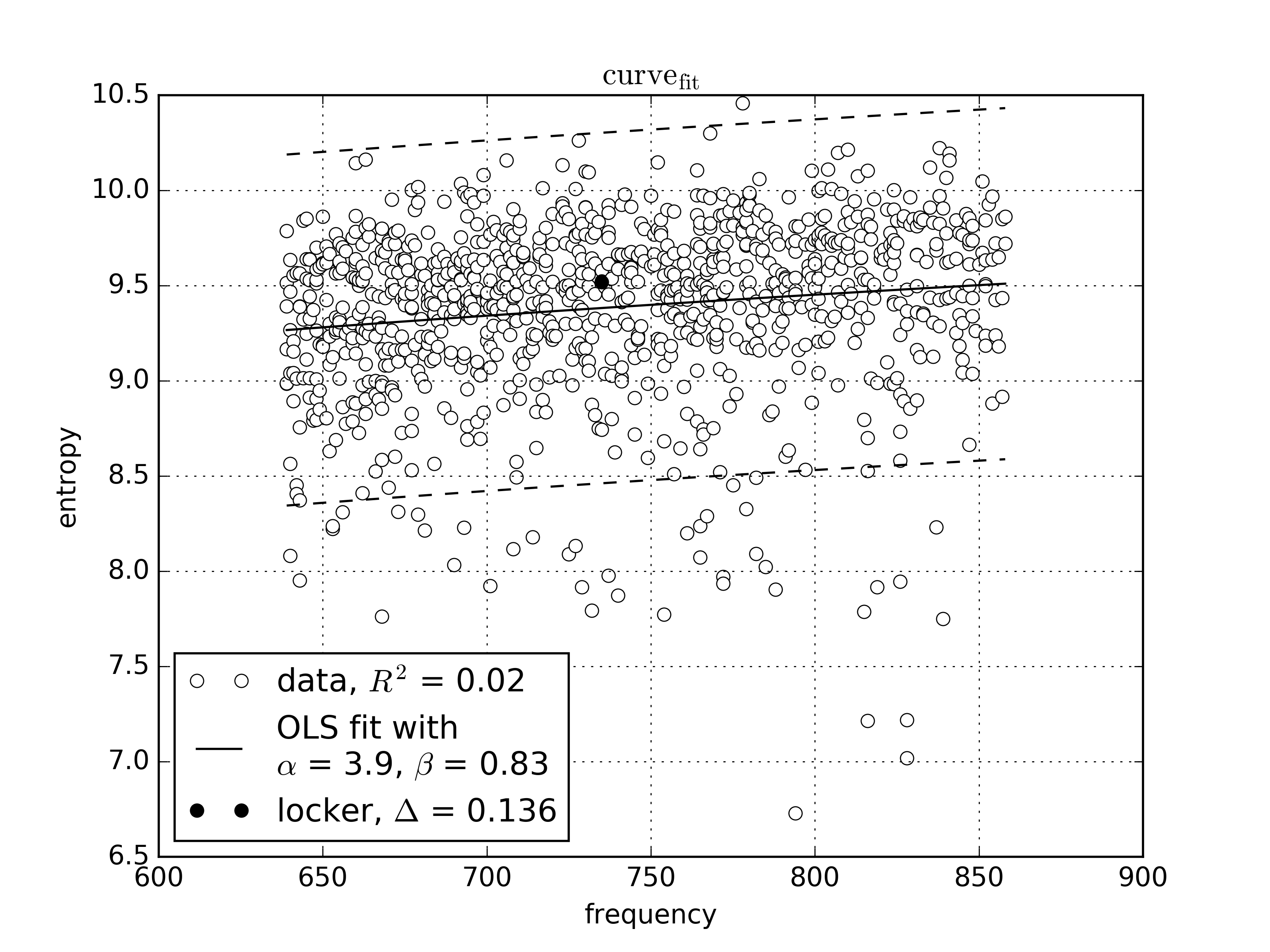}
\caption{Example of OLS for \textit{locker}}
\label{fig:regression}
\end{figure}

\paragraph{Ordinary Least Squares Regression (OLS).} Another way of normalizing entropy for frequency relies on the observation that there is a correlation between word entropy and word frequency. We try to approximate this relationship by fitting an OLS model to the observations from the corpus, where each observed word type is a data point. This approximation can then serve as a prediction for the expected change of a word's entropy given a certain change in the word's frequency. Deviations from this expectation can further be interpreted as the change in entropy solely related to semantic generality. In order to get a good approximation for each target word we only fit the model to the local $n$ data points next to the target word in the independent variable (frequency). In Figure \ref{fig:regression} we see the result of fitting the model described by Equation \ref{eq:fit} to the $1000$ data points (from a specific time period) next to the data point for the adjective \textit{locker}, `loose', in the independent variable. As we can see, the data point for \textit{locker} slightly deviates from the regression curve, more precisely, by $\Delta = 0.136$. Taking this as a starting point for the semantic development of \textit{locker} (reference time) we can now calculate \textit{locker}'s $\Delta$ in a later time period (focus time). We assume that $\Delta$ stays approximately equal if only \textit{locker}'s frequency changes. If $\Delta$, however, increases, we assume that the word underwent meaning innovation. We apply an analogous procedure to all target words.

\begin{equation}
\label{eq:fit} entropy \thicksim \alpha + \beta \ln (frequency)
\end{equation}

\subsection{Other Measures}

\paragraph{Word Frequency.} Concerning frequency, a similar argument can be brought forward as in Section \ref{sec:distributional}: When a word acquires a new meaning and can be applied to a wider range of entities, then we would expect the word to be used more often. Furthermore, it is well known that certain types of semantic change correlate with frequency. For instance, \textit{desemanticization} comes with a strong increase in frequency \citep[cf.][p.~133]{Bybee:2015aa}.
For this, we use the frequency of a word $w$ as a baseline to word entropy (parallel to the practice in hypernym detection). In order to diminish the influence of corpus size we normalize word frequency $Freq(w)$ by the number of tokens $N$ in the relevant slice of the corpus:
{\begin{equation}
Freq_n(w) = \frac{Freq(w)}{N} 
\end{equation}}

\paragraph{Second-Order Word Entropy.} A variant of word entropy used in hypernym detection is second-order word entropy where entropy is not calculated directly for the word $w$, but rather for its most-associated context words. Then the median of these is $w$'s second-order word entropy \citep[cf.][p.~40]{Santus:2014}. This measure relies on the hypothesis that the more semantically general a word is, the more it co-occurs with general context words. Presumably, this measure is more immune to the influence of word frequency, because not $w$'s own frequency plays a role, but rather the frequency of its most-associated context words. This may be helpful where we have rather accidental differences in the frequency of a word in different time periods, e.g., due to corpus size or text sort. In such a setting we reckon regular (first-order) word entropy to be more prone to these accidental factors than second-order word entropy.


\section{Diachronic Metaphor Annotation}
\label{sec:annot}

Humans often have different intuitions about what is a metaphor and what is not. According to \citet[][p.~2]{Steen10} ``the identification of metaphoric language has become a matter of controversy''. Therefore, we did not want to rely solely on our own intuitions, but identify metaphoric change of words via annotation. A number of structured annotation guidelines for synchronic metaphor identification have been proposed \citep{Group07p1, Steen10, Shutova15}. \citet[cf.][p.~8]{Steen10} distinguishes between linguistic and conceptual metaphor annotation. We adopted the former approach, since we were less interested in the exact mapping underlying a metaphoric use of a word. The crucial difference to synchronic metaphor identification is that we did not want annotators to judge individual uses but pairs of uses of lexical units.\footnote{A similar procedure is used in \citet{Erk09investigationson,Erk13} for annotation of usage similarity.} The metaphoric relation between the source and the target concept involved in the metaphoric change of a word $w$ should be reflected in $w$'s individual uses which is a common methodological assumption in historical linguistics. Individual uses bearing the meaning of source or target concept allow humans to infer these meanings which can then be judged as being (non-)metaphorical to each other. We operationalize this observation as annotation procedure.

\paragraph{Target Selection.}
We preselected the target items for annotation so that they were likely to have undergone metaphoric change. For this, we scanned the literature on metaphoric change in German such as \citet{GerdFritz98VIII} and \citet{keller2003bedeutungswandel}. The richest list we found in \citet{Paul02XXI} (ca. 140 items). However, this could not be taken directly as a gold standard. We first checked for every item whether we could attest metaphoric change in the corpus. If so, we determined a rough date of change according to when we found the metaphoric meaning clearly established in the corpus. We then checked whether the item had an occurrence frequency above a threshold of 40 around the date of change.
Only then the item was added to the test set for annotation.\footnote{We provide both: the full list of items and the one filtered for frequency.}

For every metaphoric target word $m$ in the test set we added a semantically stable word $s$ with the same POS-tag from the same frequency area. For this, we checked the words in the immediate vicinity to $m$ in the total frequency rank (of the first half of the century in which $m$'s change occurred) in DWDS, a rich online etymological dictionary of German.\footnote{\url{https://dwds.de/}} If there was no meaning change indicated and we could not attest a clear meaning change in the corpus, we added the word to the test set. Thereby, we balanced metaphoric and stable words with respect to frequency. Stable words comprise concrete words, e.g. \textit{Palast} `palace', as well as more abstract words, e.g. \textit{freundlich} `friendly'. The test set contains nouns, verbs and adjectives. (Find it in Appendix \ref{sec:testset}.)

Next, parallel to the corpus slicing (see Section \ref{sec:evaluation}), we selected 20 contexts from two time periods. These periods were set in such a way that one was located before and one after the pre-identified date of change. Supposing that a word occurs in $n$ contexts in a certain time period, we ordered them according to publication date and picked every $(n/20)$\textsuperscript{th} context guaranteeing that contexts are well-distributed over authors and the time period. Contexts with less than 10 words and obvious parsing errors were excluded in order to provide enough information for the annotators and to avoid contexts excluded by them.

Finally, contexts from the earlier period were combined randomly with contexts from the later period yielding 20 context pairs for every target. The order of every second pair was switched, minimizing the possibility that annotators infer the chronology of contexts. The pairs of all 28 target words were randomly sampled such that individual judgments were less influenced by earlier judgments of the same target, resulting in 560 context pairs presented to the annotators.

\paragraph{Annotation Procedure.} Three annotators were asked to judge for each of the 560 context pairs whether one of the contexts admitted inference of a meaning of the target word which is related metaphorically to the meaning in the other context. (Find an example in Appendix \ref{sec:guidelines}.) The annotators were linguists, two of them were marginally acquainted with historical linguistics.
The annotation guidelines are a combination and modification of the processes described by \citet{Group07p1}, \citet[][]{Steen10} and \citet{Shutova15}. Whether a meaning of a target word in context 2 (M2) is metaphorically related to the meaning in context 1 (M1) should be identified in 3 steps:

\begin{enumerate}
\setlength{\itemsep}{0pt}
\item For each word its meaning in context is established;
\item It is decided whether M1 can be seen as a more basic meaning than M2. This is the case when M2 is related to M1 in one or more of the following ways: (i), M2 is less concrete than M1; (ii), M2 is less human-oriented than M1; (iii), M2 is not related to bodily action in contrast to M1; (iv), M2 is less precise than M1.
\item If this is the case, then it is decided whether M2 contrasts with M1 but can be understood in comparison with it. If yes, M2 is judged as being metaphorically related to M1, otherwise as not being metaphorically related to M1.
\end{enumerate}
Step 2 is intended to exclude cases of non-metaphorical polysemy, for which a more basic meaning should not be identifiable \citep[cf.][p.~30]{Group07p1}. It is a rather liberal variation of the existing guidelines in that already the fact that one of the criteria holds is sufficient to consider M1 to be more basic than M2. This is because of cases like \textit{Feder}, `feather, springclip', \textit{Blatt}, `leaf, sheet, newspaper', and \textit{Haube}, `cap, cover, marriage, crest', whose meaning change would else not be captured, although we reckon it metaphoric: The change of \textit{Feder} `feather' $>$ `feather, springclip' does not fall under all criteria in step 2, e.g., there is no mapping from concrete to abstract. The existing guidelines seem to implicitly exclude such cases of metaphors, which we want to overcome. Future studies may opt for different decisions here.

Step 3 guarantees that the two meanings identified are sufficiently distinct and that there can be a mapping established between them. We cannot guarantee that annotators judge the context pairs in exactly the way we prescribe in the guidelines. (Find the full guidelines in Appendix \ref{sec:guidelines}.)

\paragraph{Annotation Results.} Annotators reported that they found the task hard, which is not surprising given that some contexts dated back 400 years making it sometimes difficult to interpret them. Accordingly, we expected this to be reflected in the inter-annotator agreement. Annotator~1 and Annotator~2 had a moderate agreement of $\kappa = .40$ (Fleiss' Kappa) for earlier and $.46$ for later contexts, while Annotator~3 had poor agreement with both, Annotator~1 ($.26$, $.26$) and Annotator~2 ($.32$, $.29$). Given this deviation, we excluded Annotator~3 from the evaluation. (Further evaluation is performed for the judgments of Annotator~1 and Annotator~2.) The agreement we found is only slightly lower than in comparable synchronic studies. \citet[][p.~21]{Group07p1}, e.g., report a $\kappa$ between 0.56 and 0.72 for different tasks. We can attribute the difference in agreement to the higher level of difficulty of the task the annotators were faced with.

The annotation results are summarized in Table \ref{tab:annotation}. Target words are ordered decreasingly according to the increase in metaphorically tagged contexts over time (last column). In addition to $\kappa$ we also give the share of items with perfect agreement ($\%$$A$), since $\kappa$ underestimates agreement on rare effects \citep{Feinstein:1990}. As you can see, the annotators overall confirmed our judgments of the targets, as most metaphoric targets are at the top of the list. Target words differ strongly in the strength of metaphoric change assigned to them: between $82\%$ (\textit{Donnerwetter}) and -$14\%$ (\textit{Haube}). Yet, most targets exhibit positive judgment, which we would expect from a test set containing metaphoric and stable targets. Striking is the position of \textit{Feder} and \textit{Haube} at the bottom, which are tagged even negatively metaphoric. This means that the share of metaphorically tagged contexts was higher for the earlier contexts. We conjecture that the reason for this is that both words were already used in other metaphoric meanings in earlier contexts. The high position of \textit{freundlich} and \textit{fett} presumably results from the fact that they are abstract adjectives. Metaphor identification for adjectives is more difficult than for nouns and verbs, because their meanings tend to be less concrete and precise \citep[cf.][p.~28]{Group07p1}. They are typically applicable to a wider range of entities, increasing the probability to encounter a context pair in our study with two uses differing in abstractness and preciseness. We will pay particular attention to the targets rated differently by us and the annotators in the analysis of the measures' predictions.

\captionsetup{belowskip=-40pt}
\begin{table}[htp]

\begin{center}

\scalebox{0.48}{

\begin{tabular}{| c | c | c | c  c  c | c | c  c  c || c |} \hline

\multirow{2}{*}{\textbf{lexeme}} & \multirow{2}{*}{\textbf{type}} & \multicolumn{4}{c|}{\textbf{earlier contexts}} & \multicolumn{4}{c||}{\textbf{later contexts}} & \multirow{2}{*}{\textbf{$\Delta$$\%$$+$}} \\  
    \cline{3-10}
	&        &	\textbf{time}	&    \textbf{$\%$$+$}    &    \textbf{$\%$$A$}  &  \textbf{$\kappa$}  &	\textbf{time} &  \textbf{$\%$$+$} & \textbf{$\%$$A$}   &    \textbf{$\kappa$}    &    \\\hline

Donnerwetter	&	met	&	1700-1800	&	.00	&	1.00	&	-	&	1850-1926	&	.82	&	.85	&	.57	&	.82	\\

peinlich	&	met	&	1600-1700	&	.00	&	.80	&	-.11	&	1800-1900	&	.67	&	.60	&	.17	&	.67	\\

gl\"anzend	&	met	&	1600-1700	&	.06	&	.85	&	.31	&	1800-1900	&	.63	&	.95	&	.89	&	.57	\\

erhaben	&	met	&	1600-1700	&	.12	&	.85	&	.49	&	1800-1900	&	.55	&	.55	&	.14	&	.43	\\

geharnischt	&	met	&	1700-1800	&	.00	&	.95	&	-.03	&	1850-1926	&	.42	&	.95	&	.90	&	.42	\\

freundlich	&	sta	&	1600-1700	&	.07	&	.70	&	.10	&	1800-1900	&	.38	&	.65	&	.35	&	.31	\\

fett	&	sta	&	1600-1700	&	.08	&	.65	&	.06	&	1800-1900	&	.27	&	.55	&	.17	&	.20	\\

flott	&	met	&	1700-1800	&	.00	&	.85	&	.72	&	1850-1926	&	.20	&	.75	&	.59	&	.20	\\

Blatt	&	met	&	1500-1600	&	.00	&	.75	&	-.10	&	1700-1800	&	.17	&	.60	&	.16	&	.17	\\

Rausch	&	met	&	1600-1700	&	.00	&	.85	&	.50	&	1800-1900	&	.15	&	.65	&	.36	&	.15	\\

locker	&	met	&	1700-1800	&	.11	&	.90	&	.70	&	1850-1926	&	.23	&	.65	&	.30	&	.12	\\

ausstechen	&	met	&	1600-1700	&	.10	&	1.00	&	1.00	&	1800-1900	&	.21	&	.95	&	.86	&	.11	\\

eitel	&	met	&	1600-1700	&	.00	&	.35	&	-.27	&	1800-1900	&	.11	&	.45	&	-.07	&	.11	\\

ahnen	&	sta	&	1600-1700	&	.00	&	.70	&	.20	&	1800-1900	&	.09	&	.55	&	.11	&	.09	\\

br\"uten	&	met	&	1600-1700	&	.11	&	.90	&	.66	&	1800-1900	&	.19	&	.80	&	.48	&	.08	\\

erdenklich	&	sta	&	1700-1800	&	.00	&	.60	&	-.25	&	1850-1926	&	.06	&	.80	&	.22	&	.06	\\

aufwecken	&	sta	&	1600-1700	&	.24	&	.85	&	.62	&	1800-1900	&	.27	&	.75	&	.43	&	.03	\\

stillschweigend	&	sta	&	1700-1800	&	.07	&	.75	&	.13	&	1850-1926	&	.08	&	.60	&	-.07	&	.02	\\

bewachsen	&	sta	&	1700-1800	&	.00	&	.85	&	-.08	&	1850-1926	&	.00	&	1.00	&	-	&	.00	\\

Palast	&	sta	&	1700-1800	&	.00	&	.80	&	-.11	&	1850-1926	&	.00	&	.80	&	-.11	&	.00	\\

Fenchel	&	sta	&	1600-1700	&	.00	&	.95	&	-.03	&	1800-1900	&	.00	&	1.00	&	-	&	.00	\\

Wohngeb\"aude	&	sta	&	1700-1800	&	.00	&	.95	&	-.03	&	1850-1926	&	.00	&	1.00	&	-	&	.00	\\

adelig	&	sta	&	1600-1700	&	.08	&	.65	&	.11	&	1800-1900	&	.07	&	.70	&	.10	&	-.01	\\

Evangelium	&	sta	&	1500-1600	&	.05	&	.95	&	.64	&	1700-1800	&	.00	&	.90	&	-.05	&	-.05	\\

Unh\"oflichkeit	&	sta	&	1600-1700	&	.05	&	.95	&	.64	&	1800-1900	&	.00	&	.65	&	-.21	&	-.05	\\

heil	&	sta	&	1600-1700	&	.13	&	.40	&	-.01	&	1800-1900	&	.00	&	.50	&	.03	&	-.13	\\

Feder	&	met	&	1700-1800	&	.28	&	.90	&	.76	&	1850-1926	&	.13	&	.75	&	.28	&	-.14	\\

Haube	&	met	&	1600-1700	&	.20	&	.75	&	.37	&	1800-1900	&	.06	&	.90	&	.44	&	-.14	\\
\hline
all	&	-	&	-	&	.06	&	.80	&	.40	&	-	&	.20	&	.74	&	.46	&	.14	\\\hline

\end{tabular}}

\caption{\small Annotation results divided into judgments for earlier and later contexts. $\%$$+$ contains the share of metaphorically tagged items in all items for the respective target word on which there was perfect agreement. $\%$$A$ gives the share of items with perfect agreement and $\kappa$ the Fleiss' Kappa score for all annotators. The last column, $\Delta$$\%$$+$, contains the relative increase or decrease in metaphorically tagged items over time calculated by $(\%+_{later}) - (\%+_{earlier})$. Rows are ordered decreasingly according to the values in $\Delta$$\%$$+$.}
\label{tab:annotation} 
\end{center}

\end{table}

\captionsetup{belowskip=-20pt}


\section{Evaluation}
\label{sec:evaluation}

As with \citet{Gulordava11} or \citet{Hamilton:2016}, we assess the measures' performance by comparing their predictions in a corpus against a gold standard. Our gold standard is the rank of target words in Table \ref{tab:annotation} obtained by annotation. We obtain the measures' predictions for the target words by first calculating their values in a time period 1 before the starting point of change and in a time period 2 after that. We then compute the difference $d$ in values between period 1 and 2 for each target word and further rank the target words according to $d$. Next, we compute the rank correlation between each of these predicted ranks and the gold rank as a performance measure.

Time period 1 is usually the century before and period 2 the century after the century of change, e.g., \textit{ausstechen} (1739) will be compared in 1600-1700 and 1800-1900. (Only for targets from 1800-1900 time period 2 is different, i.e., 1850-1926, since the corpus version we use only contains texts until 1926.) Stable words are compared in the same time periods as their metaphoric counterparts (see Section \ref{sec:annot}). With this procedure we have the possibility to evaluate the measures (i), only on targets from the same century, fixing influential side factors such as corpus size, and (ii), on all targets, which is a much harder task. (Find a list of time periods with corpus sizes in Appendix \ref{sec:params}.)


\section{Results}
\label{sec:Results}

Table \ref{tab:resultssum} shows Spearman's $\rho$ quantifying the correlation between the measures' predicted ranks and the gold standard rank. We can directly see that word entropy (H) correlates significantly with the gold rank in different conditions. Moreover, the ranking it predicts for targets from 1700-1800 correlates much stronger (.64) with the gold rank than the other measures' predictions. Note that the correlation is highly significant despite the relatively small sample size. In the harder condition, where we look at the ranks across different time periods, H still correlates significantly and stronger than all other measures with the gold rank. However, apart from H, the conclusions we can draw about the other measures can only be preliminary, as there is no significance for their predicted ranks.

At first glance, the normalized versions of entropy do not perform as expected: H\textsubscript{{\tiny MON}} never outperforms frequency and shows even negative correlation in one time period. Since we reckoned that the reason for this is the low setting of the hyperparameter $n = 29$ (which we adopted with the intention to construct all vectors from a common number of contexts), we also tested the measure on target words from 1700-1800 with a setting of $n$ such that the maximum number of contexts is used to construct the word vector and the number of vectors to average over $k = 10$. In this setting H\textsubscript{{\tiny MON}}'s prediction has a highly significant correlation with the gold rank which is comparable in strength to H.

Notably, H\textsubscript{{\tiny OLS}} has the best performance for targets from 1800-1900. We tried out different hyperparameter settings and found that our initial choice of the data window size $n = 1000$ may also not have been optimal, as higher $n$ yield better, yet non-significant, results: $n =$ 500/10000/20000/50000 yields $\rho =$ 0.19/0.32/0.31/0.21 respectively, for targets from 1700-1800. Another factor possibly biasing H\textsubscript{{\tiny OLS}} are different variances in different corpora or frequency areas which may also connect to our observation that the measure correlates negatively with absolute changes in frequency, i.e., decrease in frequency often leads to increase in H\textsubscript{{\tiny OLS}} and vice versa.

H\textsuperscript{2} consistently performs poorly. Moreover, testing of different values for $N$ yields a wide range of $\rho$ values between -0.29 and 0.42 for targets from 1700-1800, not allowing conclusions on the performance of the measure because the correlation is not significant.

Analyzing the predicted ranks reveals interesting insights. H and its normalized siblings rank \textit{Donnerwetter}, which is at the top of the gold rank, at the very bottom. This is, presumably, because in its later metaphoric sense `blowup' the word can be used as an interjection in very short sentences as in (\ref{ex:donner}).

\begin{examples}
\item \label{ex:donner} {\em Potz Donnerwetter!} \footnote{Hauptmann, Gerhart: Der Biberpelz. Berlin, 1893.}
      \glt `Man alive!'
\end{examples}
This narrows down \textit{Donnerwetter}'s contextual distribution due to our model only considering words within a sentence as context. H\textsuperscript{2} and frequency are not sensitive to this and rank the word much higher. This shows that, (i), different factors play a role in determining the contextual distribution of a word suggesting that a model of semantic change should incorporate different types of information and, (ii), that H\textsuperscript{2} and frequency may still be helpful in detecting metaphoric change in certain settings. The dominance of H may also be a hint to this direction: Word entropy combines frequency and contextual distribution as it is influenced by both.

\textit{Feder} and \textit{Haube} from the very bottom of the gold rank are not beyond the bottom-items of any measure's prediction. In H's prediction, which is the best-performing measure, they rank near the middle (12, 18). This indicates that their position at the bottom of the gold rank may not accurately reflect the semantic change they underwent. Similarly for the adjectives \textit{freundlich} and \textit{fett} ranking in all predictions near middle or lower (for H: 18, 10). We still have to assess how these words behave in future studies.

\begin{table}[htp]
\begin{center}
\scalebox{0.8}{
\begin{tabular}{l|ccc|c}

							&	\textbf{1600-1700}	&	\textbf{1700-1800}	&	\textbf{1800-1900}	&	\textbf{all} 	\\\hline
\textbf{H} 						&	1.00	&	\textbf{.64***}	&	 .10	&	\textbf{.39*}		\\
\textbf{H\textsubscript{{\tiny MON}}}	&	1.00	&	.19	&	-.10	&	.06	\\
\textbf{H\textsubscript{{\tiny OLS}}}	&	1.00	&	.20	&	.29	&	.26	\\
\textbf{H\textsuperscript{2}}		&	1.00	&	.06	&	.02 	&	.13	\\\hline
\textbf{Freq\textsubscript{n}}					&	1.00	&	.29	&	-.07 	&	.26	\\

\end{tabular} }
\caption{\small Summary of the predictions of word entropy (H), H normalized via MON (H\textsubscript{{\tiny MON}}), H normalized via OLS (H\textsubscript{{\tiny OLS}}), second-order word entropy (H\textsuperscript{2}) and normalized frequency (Freq\textsubscript{n}) for the respective subset of target words from our test set for each time period. Values in cells refer to Spearman's rank correlation coefficient $\rho$ between the individual measure's predicted rank and the relevant subrank from the annotated gold standard (Table \ref{tab:annotation}).}
\label{tab:resultssum}
\end{center}
\end{table}


\section{Conclusion}
\label{sec:Conclusion}

Semantic generality is an important indicator of semantic change. As \citet[cf.][p.~197]{Bybee:2015aa} puts it, generalization and specialization are two basic principles of meaning change. We proposed a way to detect metaphoric change based on semantic generality and built a test set for the evaluation of computational models of metaphoric change in German. We proposed an annotation procedure strictly derived from comparable synchronic work and showed that annotators can show reasonable agreement. Different distributional measures based on the information-theoretic concept of entropy were compared against the annotators judgments and it was found that raw word entropy correlates strongly and significantly with the gold rank in different settings in contrast to most other entropy measures and frequency. We found evidence that H\textsubscript{{\tiny MON}} predicts well with certain parameter settings.

Both, the annotation procedure and the computational model, are generalizable to different types of semantic change. Moreover, our model is unsupervised and language-independent as it relies, in principle, on minimal linguistic input, since entropy can be computed already from a raw token co-occurrence matrix. Yet, the model profits from richer input as indicated in \citet[][]{Shwartz16}.

Future studies should test how well word entropy distinguishes metaphoric change from other types of meaning innovation and how well it detects innovative and reductive meaning change in general. The latter may be tested straightforwardly on the English data of \citet{Gulordava11} and \citet{Hamilton:2016}. In doing so, it will be interesting to see how our model performs in comparison to diachronic similarity and WSI models.


\section*{Acknowledgments}

We thank Prof. Dr. Sebastian Pad{\'o} (Institute for Natural Language Processing, University of Stuttgart) for pointing out his idea to normalize word entropy via OLS. We are very grateful to Prof. Dr. Olav Hackstein (Historical and Indo-European Linguistics, LMU Munich) and his research colloquium for valuable discussions and comments and to Sascha Schlechtweg for statistical advice. We would like to thank Andrew Wigman for careful proof-reading as well as J{\"o}rg F{\"o}rstner, Michael Frotscher, Altina Mujkic, Edona Neziri, Cornelia van Scherpenberg, Christian Soetebier and Veronika Vasileva for help related to the annotation process. Last but not least, we thank the reviewers for constructive criticism helping us to improve the paper substantially.


\bibliographystyle{aclnatbib}
\bibliography{/Users/admin/Documents/workspace-python/backup/literature/Bibliography-general.bib}



\appendix

\section{Hyperparameters and Corpus Preprocessing Details}
\label{sec:params}

\subsection{Hyperparameters}
Second-order word entropy has 3 hyperparameters: (i), the number of positively associated contexts $N$ to compute the average/median from; (ii), whether to use median or average entropy among the top $N$ contexts;\footnote{Note that for any test pair, $N$ is the \emph{maximal} number of associated contexts, which is reduced to $M$ if a test target has only $M$ ($<N$) positively associated contexts in one of the two matrices to compare.} and (iii), the association metric used to sort the contexts by relevance (i.e., PPMI or PLMI). We choose the following combination of hyperparameters: $\langle \text{100, median, PLMI} \rangle$, which is suggested by the work of \citet{Shwartz16}.

For MON entropy normalization we choose $n = 29$, because that is the lowest context number of a word in one of its two relevant time periods, and $k = 10000$. For OLS normalization we choose $n = 1000$.

\subsection{Corpus Preprocessing}

Words that occur less than 5 times in the whole corpus, functional words and punctuation are deleted. As functional words we regard those which are not tagged with a POS-tag starting with either `N', `V' or `AD'. Every token is then replaced by its lemma form combined with the starting of its POS-tag, e.g., \textit{geht} is replaced by \textit{gehen:V}. Note that both diachronic lemmatization and POS-tagging are provided by DTA.

\begin{table}[htp]
\begin{center}
\tabcolsep=0.11cm
\scalebox{0.65}{
\begin{tabular}{c|ccccc}

\textbf{time period} & 1500-1600 & 1600-1700 & 1700-1800 & 1800-1900 & 1850-1926 	\\\hline
\textbf{corpus size} &	0.2M & 13M & 23M & 34M & 23M \\

\end{tabular} }
\caption{Time periods for evaluation and their respective corpus sizes after preprocessing.}
\label{tab:corpus}
\end{center}
\end{table}


\section{Annotation Guidelines}
\label{sec:guidelines}

\subsection{Introduction}

Following the terminology from \citet[cf.][p.~24]{Koch:2016} \textit{innovative meaning change}, as opposed to \textit{reductive meaning change}, is where the existing meaning $M_A$ of a word acquires a new meaning $M_B$, where this normally happens over a long period of time.

	\begin{description}

	\item[Metaphoric Change] is, then, a subcategory of innovative meaning change (besides metonymic change, generalization...) where $M_B$ is related to $M_A$ by similarity or a reduced comparison \citet[cf.][p.~47]{Koch:2016}. \citep[cf.~also][p.~10]{Steen10}

	\end{description}	
Note that the annotation process described below is a combination and modification of the processes described by \citet{Group07p1}, \citet[][]{Steen10} and \citet{Shutova15}.


\subsection{Annotation Process}
\label{section:annotation}

You will be given an OpenOffice table document with approximately 560 lines. In every line you will see in columns 2 and 3 two uses of a word (the target word contained in column 1) with its surrounding contexts. The relevant word is marked in bold font in both contexts.

\begin{enumerate}

	\item For each such use of a word establish its meaning in context, that is, how it applies to an entity, relation, or attribute in the situation evoked by the text (contextual meaning). Take into account what comes before and after the word. Note that the word might be used differently from what you are familiar with. Don't let yourself be confused by alternative spelling.

	\item Try to find an interpretation where the meaning in the second context (M2) is related to the meaning 
in the first context (M1) in one or more of the following ways:

	\begin{itemize}

		\item M2 is less concrete than M1 (what it evokes is harder to imagine, see, hear, feel, smell, and taste);

		\item M2 is less human-oriented than M1;

		\item M2 is not related to bodily action in contrast to M1;

		\item M2 is less precise than M1 (precise as opposed to vague).

	\end{itemize}

\item If M2 is indeed related to M1 in one or more of these ways, decide whether M2 contrasts with M1 but can be 
understood in comparison with it. (See below for an example.)

	\item \begin{enumerate}[(i)]

	\item If yes, \underline{fill in 1} into the column headed by `M2 is metaphorically related to M1', judging M2 as \underline{being metaphorically related} to M1.
	
	\item If no, \underline{fill in 0} into the column headed by `M2 is is metaphorically related to M1', judging M2 as \underline{not being metaphorically related} to M1.

	\item If you cannot decide, e.g., because the word marked in bold font doesn't match the word shown in column 1 in meaning or part of speech, you don't understand either of the contexts, one is too unspecific or other reasons, don't perform evaluation, \underline{fill in a 1 into the comments column} and go on to the next test item.

	\end{enumerate}

\item Compare the two meanings in the other direction, i.e., decide whether M1 is metaphorically related to M2 by going through steps 2 to 4 and fill your judgment into the column 
headed by `M1 is metaphoric compared to M2'.

\end{enumerate}

Please make sure that you don't change anything in the file apart from column width, your judgments and comments. Finally, return the annotated document to the above-mentioned email address. If you have any further questions on the task, don't hesitate to ask.


\subsection{Annotation Example}

The following example illustrates how the procedure operates in practice. Consider Table \ref{tab:example} as an example table similar to the one you will receive for annotation.

\captionsetup{belowskip=-17pt}

\begin{table*}[htp]

\begin{center}

\tabcolsep=0.11cm

\scalebox{0.77}{

\begin{tabular}{| p{2.0cm} | p{5cm} | p{5cm} | p{3cm} | p{3cm} | c |} \hline
target word & Context 1 & Context 2 & M1 is metaphorically related to M2 & M2 is metaphorically related to M1 & comments \\\hline
umw\"alzen &

Ein Knecht vnd Tagel\"ohner hat doch auff den abendt sein Brodt / lohn vnd ruhe / Ein Kriegsman seinen Monat soldt / ich aber mus der elenden n\"achte viel haben / da mich mein au{\ss}wendiger schmertz vnd inwendige hertzen angst nit schlaffen lest / ja ich bin der elendeste Mensch auff Erden / wann andere Leute / auch das tumme Vieh in jhrem Stalle jhre leibliche bequeme nachtruhe haben / mu{\ss} ich mich \textbf{vmbweltzen}/ vnd kan keinen schlaff in meine augen bringen & 

Kinadon wollte den Staat \textbf{umw\"alzen}, weil er nicht, obgleich von starkem und th\"aigem Geiste, zu den Gleichen geh\"orte.
	
& 0 & 1 & \\\hline

umw\"alzen & Bey diesen und \"ahnlichen Handlungen ist das Auge entweder offen, oder verschlossen, aber in beyden F\"allen der Augapfel krampfhaft \textbf{umgew\"alzt}, so dass nur der Rand der Iris unter dem obern Augenliede hervorscheint, die Pupille erweitert, und die Netzhaut unempfindlich selbst gegen die heftigsten Reitzmittel. &

Und was sagestu? habe ich deinen so hochger\"uhmten Ritter dann auch vom Pferde gezaubert / da er sich im Sande \textbf{umweltzete}?

& 0 & 0 & \\\hline

... & & & & & \\\hline

\end{tabular}}

\caption{\label{tab:example} Example Annotation Table}
\end{center}

\end{table*}

In line 1 you need to compare two uses of the word \textit{umw\"alzen}. In context 1 \textit{umw\"alzen} is used in  the sense `to turn around something or someone physically' (M1). This contrasts with its use in context 2 where it is used in the sense `to change something radically' (M2). M2 is clearly less concrete than M1 and not necessarily related to bodily action. Moreover, M2 is less precise, since we may have greater disagreement about the question whether something `changed radically' than we may have on the question whether someone or something (was) turned around. (You may have a different intuition here, which should then be reflected in your judgment accordingly.)

Now, as we saw, M2 contrasts with M1. However, it can be understood in comparison with it: We can understand abstract change in terms of physical or local change. Consequently, we fill in 1 in the column headed by `M2 is metaphorically related to M1', judging M2 to be metaphorically related to M1. 
And, for the same reasons as mentioned above, we fill in a 0 in the column headed by `M1 is metaphorically related to M2'.

In line 2 both meanings of \textit{umw\"alzen}, M1 and M2, are similarly concrete, human-oriented, related to bodily action and precise. They don't contrast with each other. (You may want to say that they are equal.) Hence, neither meaning has a metaphoric relation to the other. Consequently, we fill in 0 into both columns.



\newpage
~
\newpage

\section{Metaphoric Change Test Set}
\label{sec:testset}

\begin{table}[!ht]
\begin{center}
\tabcolsep=0.11cm
\scalebox{0.6}{
\begin{tabular}{|l|l|l|p{7cm}|l|l|}
\hline 
\bf lexeme & \bf POS & \bf type & \bf old meaning $>$ new meaning & \bf date & \bf freq. \\ 
\hline
eitel & AD & met & `empty' $>$ `arrogant' & 1764 & 1320\\
freundlich & AD & sta & `cordial, benevolent' & 1516 &  1351 \\
\hline
erhaben & AD & met & `pronounced, prominent' $>$ `distinguished, great' & 1725 & 1003 \\
fett & AD & sta & `obese, greasy, fatty (food)' & 1557 & 951 \\
\hline
gl\"anzend & AD & met & `sparkling, luminous' $>$ `sparkling, luminous, very good' & 1753 & 496 \\
adelig & AD & sta & `aristocratic, noble, virtuous' & 1585 & 481 \\
\hline
peinlich & AD & met & `painful' $>$ `painful, diligent, embarrassing' & 1788 & 440 \\
heil & AD & sta & `safe, sound' & 1494 & 423 \\
\hline
locker & AD&  met & `not tense/tight' $>$ `frivolous, loose' & 1800 & 407 \\
stillschweigend & AD & sta & `silent' & 1603 & 498 \\
\hline
geharnischt & AD & met & `armoured' $>$ `sharply-worded, strong' & 1825 & 50 \\
bewachsen & AD & sta & `overgrown'  & 1603 & 52 \\
\hline
flott & AD & met & `afloat' $>$ `lively, quick, dressy' & 1800 & 42 \\
erdenklich & AD & sta & `imaginable' & 1647 & 44 \\
\hline \hline
Feder & N & met & `feather' $>$ `feather, springclip' & 1852 & 1121 \\
Palast & N & sta & `palace, chateau' & 1500 & 1111 \\
\hline
Blatt & N & met & `leaf' $>$ `leaf, sheet, newspaper' & 1638 & 410 \\
Evangelium & N & sta & `the Gospel' & 1521 & 405 \\
\hline
Haube & N & met & `cap' $>$ `cap, cover, marriage, crest' &  1712 & 138 \\
Fenchel & N & sta & `fennel' & 1531 & 138 \\
\hline
Rausch & N & met & `intoxication (due to use of mind-altering substances)' $>$ `inebriation' & 1756 & 65 \\
Unh\"oflichkeit & N & sta & `discourtesy' & 1605 & 65 \\
\hline
Donnerwetter & N & met & `thunderstorm' $>$ `thunderstorm, blowup' & 1805 & 49 \\
Wohngeb\"aude & N & sta & `residential building' & 1737 & 49 \\
\hline \hline
br\"uten & V & met & `breed' $>$ `breed, brood over sth.' & 1754 & 184 \\
aufwecken & V & sta & `wake up (so.)' & 1585 & 183 \\
\hline
ausstechen & V & met & `excise' $>$ `excise, outrival' & 1739 & 53 \\
ahnen & V & sta & `suspect' & 1500 & 53 \\
\hline
\end{tabular} }
\end{center}
\caption{\small Historical data: sta (stable), met (metaphoric). The date column indicates the year of the occurrence of the change for metaphoric items, but the year of the first occurrence for stable items. The last column (freq.) lists the frequency of the lexeme in the first half of the century in which the corresponding metaphoric change occurs.} 
\label{tab:testset}
\end{table}

\end{document}